\documentclass[sigconf,authorversion,nonacm]{acmart}
\usepackage{graphicx} 
\usepackage{listings}
\usepackage{adjustbox}
\usepackage{xspace}
\usepackage[font={small}]{caption}

\def\ourmodel{LLMD\xspace}

\title{\ourmodel: A Large Language Model for Interpreting \\Longitudinal Medical Records}

\author{Robert Porter, Adam Diehl, Benjamin Pastel, J. Henry Hinnefeld, Lawson Nerenberg,\\Pye Maung, Sebastien Kerbrat, Gillian Hanson, Troy Astorino, Stephen J. Tarsa}
\affiliation{%
  \institution{PicnicHealth}
  \city{San Francisco}
  \state{CA}
  \country{USA}
}

\date{October, 2024}
\begin{abstract}
We introduce \ourmodel\footnote{LLMD is a portmanteau of LLM (Large Language Model) and MD (Medical Doctor)}, a large language model (LLM) designed to analyze a patient's medical history based on their medical records. Along with domain knowledge,
\ourmodel is trained on a large corpus of records collected over time and across facilities, as well as tasks and labels that make nuanced connections among them. This approach is critical to an accurate picture of patient health, and has distinctive advantages over models trained on knowledge alone, unlabeled records, structured data from electronic health record (EHR) aggregators, or records from a single health system. Today, \ourmodel is deployed to support virtual care, care coordination, and the curation of datasets behind 60+ research studies, including data submitted to the FDA.

 The recipe for \ourmodel first continues pretraining a foundational model on both domain knowledge and the contents of millions of records. These span an average of 10 years of care and as many as 140 care sites per patient. \ourmodel is then instruction fine-tuned on \emph{structuring} and \emph{abstraction} tasks -- the former jointly identify and normalize document metadata, provenance information, clinical named-entities, and ontology mappings, while the latter roll these into higher-level representations, such a continuous era of time a patient was on a medication. \ourmodel is deployed within a layered validation system that includes continual random audits and configurable review by experts, e.g. based on uncertainty, disease-specific rules, or end use-case. This provides feedback to improve \ourmodel and fine-grained control over data quality for a spectrum of needs.

\ourmodel exhibits large gains over both more-powerful generalized models and domain-specific models. On medical knowledge benchmarks, \ourmodel-8B achieves state of the art accuracy on PubMedQA text responses, besting orders-of-magnitude larger models. 
On production tasks, we show that LLMD significantly outperforms all other models evaluated, and among alternatives, large general purpose LLMs like GPT-4o are more accurate than models emphasizing medical knowledge. We find strong evidence that accuracy on today's medical benchmarks is not the most significant factor when analyzing real-world patient data, an insight that validates our approach and has implications for future medical LLMs.

\end{abstract}

\begin{document}
\maketitle
\section{Introduction}
\label{sec:intro}

\begin{figure}[t]
\centering
\includegraphics[width=0.475\textwidth]{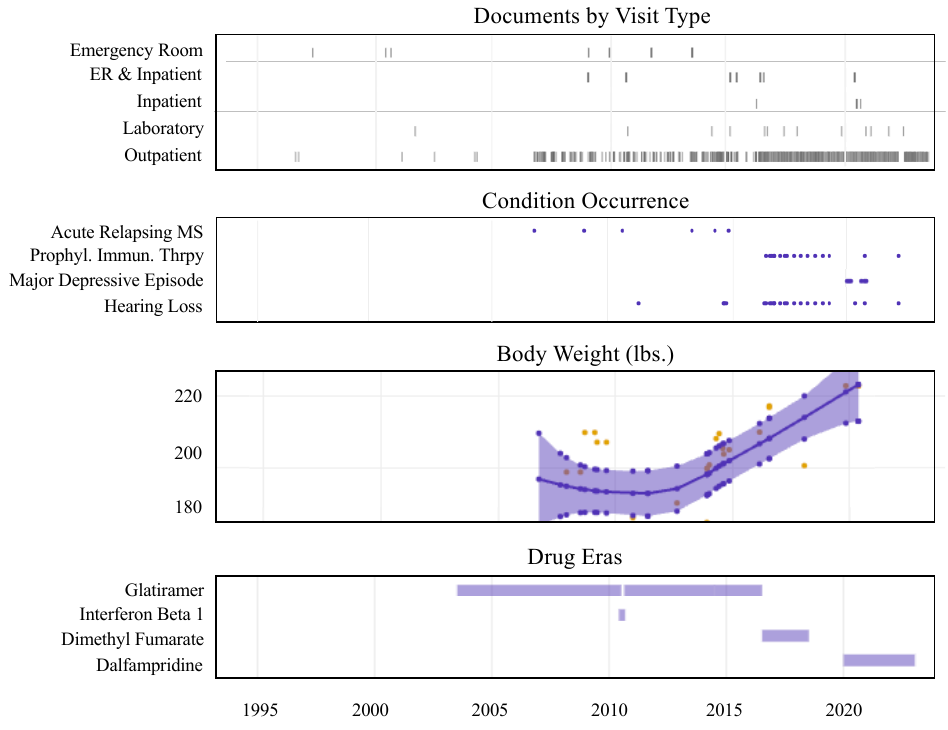}
\caption{Data from a patient with Multiple Sclerosis (MS). Their health information spans 27 years of care, 705 visits to 40 specialist providers, and 5 health systems in 2 states. Once longitundinal records are structured, we can abstract measurements and drug eras to intuitively model this patient's care journey.}
\label{fig:abstracted_data}
\end{figure}

\ourmodel is a large language model (LLM) for understanding and analyzing a patient’s medical history. Today, it is deployed to power patient- and research-facing products offered by PicnicHealth. LLMs represent an astonishing breakthrough in Artificial Intelligence (AI)~\cite{chang2024survey, OpenBioLLMs, touvron2023llama}, and exhibit nuanced pattern matching and information recall capabilities. In the medical domain, LLMs fine-tuned on domain knowledge can appropriately respond to licensing exam questions and patient queries~\cite{brin2023large, singhal2023large,OpenBioLLMs,Nori2023Capabilities}. Techniques like Retrieval Augmented Generation (RAG) and Chain of Thought (CoT) suggest paths to deployment in applications that demand trustworthy outputs, e.g. by citing evidence or explaining responses~\cite{lewis2020retrieval, wei2022chain}. Building on this promise, this paper presents an LLM designed for use cases requiring a detailed understanding of patient health at the highest accuracy standards.

Today, narrative text in medical records is the richest source of patient health information available. However, tapping into it remains difficult due to important patient privacy concerns, technical barriers to access, and the difficulty of modeling record contents~\cite{Sherman2016Real-World}. Compounding these issues, longitudinality matters: information captured in records over time and across facilities is critical to an accurate picture of patient health. As one example, in a dataset of several thousand hemophilia patients, we found contradictory diagnoses of disease subtype in the records of 30\% of patients. For them, an LLM answering even simple questions about their primary condition requires access to several records along with the intelligence to weigh information based on its provenance and connections to other evidence. 

\ourmodel does just that, based on a training dataset that combines medical knowledge with labeled longitudinal records from the PicnicHealth platform. This platform powers patient-centric tools for virtual care, care coordination, and therapy development, in part by retrieving and managing records on behalf of patients. Data curated in collaboration with patients includes millions of records, collected from 100k+ sites, which span decades of care for most patients; ultimately, \ourmodel has access to 350M labels from more than 10 years of labeling by human \emph{clinical data abstractors}~\cite{PHWebsite}.

\ourmodel's recipe combines continued pretraining of \texttt{llama3.1} ~\cite{touvron2023llama, cui2024efficienteffectivetextencoding, roziere2024codellamaopenfoundation} with task-driven instruction fine-tuning ~\cite{Wei2021Finetuned}. Continued pre-training (Section~\ref{sec:pretrain}) teaches the LLM the patterns and statistics of both electronic and paper records, as well as general medical knowledge. \emph{Structuring} tasks (Section~\ref{sec:structuring}) teach \ourmodel to produce normalized data from arbitrary record contents, while \emph{abstraction} tasks (Section~\ref{sec:abstraction}) mimic clinicians to capture the clinical view of patient health. We find that only once records are both structured \emph{and} abstracted can we draw insights suitable for real-world use cases
 (Figure~\ref{fig:abstracted_data}). 

In production, \ourmodel's outputs are subject to multiple layers of validation to ensure consistency and accuracy. These include secondary models that predict performance relative to abstractors performing the same task, rule-based data conformance and plausibility checks, and manual auditing by abstractors and clinicians. Outputs failing at any layer are corrected or suppressed and folded into future training data. These mechanisms are configurable based on use case, allowing us to process low risk data quickly and efficiently when appropriate, or guarantee that a abstractor verifies data using protocols acceptable to regulators when needed. 

We evaluate LLMs including \ourmodel on both common medical benchmarks and on production tasks, developing strong evidence that \emph{LLMs must be trained on complete, labeled longitudinal records to accurately model patient health and care}.
In terms of benchmark performance, shaping our continued-pretraining process towards the PubMedQA benchmark leads \ourmodel-8B to achieve state-of-the-art text responses, beating both general and domain-tailored models with much larger parameter count. But we observe that many models struggle to leverage medical knowledge when they must contend with messy real-world records. This effect is most pronounced when we analyze performance on production structuring and abstraction tasks. We show that \ourmodel significantly outperforms all other models evaluated and that among alternatives, large general purpose LLMs far outperform those emphasizing medical knowledge. This dynamic -- that accuracy working with records is only partly driven by an LLM's medical knowledge -- validates the importance we place on both pre-training and detailed instruction fine-tuning on real records, while demonstrating an important gap facing LLMs developed for real-world medical uses.

This paper proceeds as follows: Section 2 describes the PicnicHealth platform, our source for training data as well as  our target environment for deployment. Section 3 introduces our continued pretraining and instruction fine-tuning approaches, while describing our training dataset and context generation procedures. Section 4 discusses validation mechanisms. Section 5 evaluates \ourmodel against alternatives on common benchmarks, tasks pulled from our production data, and investigates performance on infrequent but clinically important long-tail concepts.

\section{Medical Records on the PicnicHealth Platform}
\label{sec:platform}

PicnicHealth works with patients to retrieve and manage their medical records, regardless of the format they are in or the facility holding them. This includes electronic records as well as paper-based records. Paper records are of particular importance for visits to providers practicing outside of large health systems, for historical records produced before Electronic Health Records (EHRs) were ubiquitous, and for facilities whose systems impede sharing. PicnicHealth's platform is format-agnostic under its base case support for processing physical copies of records. It is also facility agnostic based on the legal right patients have to access records on request. 

By building a complete, longitudinal picture of one's health, PicnicHealth is able to offer compelling services to both patients and researchers. For patients, PicnicHealth offers products for virtual care, care coordination, and records-management. For researchers, PicnicHealth offers products to improve the speed, flexibility, and cost of observational studies supporting new therapies by directly engaging with willing patients. 

\section{Tasks \& Training Data}
\label{sec:tasks}
This section describes how we build \ourmodel, including 
the details of our tasks and training data.

\subsection{Continued Pre-Training}
\label{sec:pretrain}
Table~\ref{tab:pretraining} shows the breakdown of the dataset we use to continue pretraining a foundational model using a large unlabeled corpus. This step follows the same training configuration as the foundational model, e.g. next-token prediction for Llama models. The primary purpose is to adapt the foundational model to the patterns of medical records, while imbuing it with key information needed by downstream tasks~\cite{cui2024efficienteffectivetextencoding, roziere2024codellamaopenfoundation}. Our dataset consists of 28B tokens, approximately $90\%$ sourced from records and $10\%$ representing medical knowledge. Along with widely-available sources for general medical knowledge, we include two data sources supporting PubMedQA -- the PubMedQA training dataset enriched with CoT, and papers from PubMed Central -- but none directly supporting other benchmarks. This deliberate choice shapes \ourmodel to PubMedQA for the results in this paper, allowing us to analyze how well knowledge transfers to other benchmarks and to map out the relationship between accuracy on benchmarks and records-processing. 

\begin{table}
    \centering
    \begin{tabular}{lcc}
         \textbf{Corpus}
         &  \textbf{
                \begin{tabular}[c]{@{}c@{}}
                    Token \\ Count
                \end{tabular}
            }
         & \textbf{
                \begin{tabular}[c]{@{}c@{}}
                    \% Training \\ Samples
                \end{tabular}
            } \\
    \cline{1-3}
    \cline{1-3}
         \multicolumn{1}{l|}{Electronic Records}&  16B& 59.5\%\\
         \multicolumn{1}{l|}{Paper Records}&  9B& 32.3\%\\
         \multicolumn{1}{l|}{PubMedQA}&  758M& 2.7\%\\
         \multicolumn{1}{l|}{PubMed PMC (Journal Articles)}&  290M& 1.0\%\\
         \multicolumn{1}{l|}{Ontologies}&  49M& 0.2\%\\
         \multicolumn{1}{l|}{Medical Knowledge Resources}&  1.2B& 4.4\%\\
    \end{tabular}
    \caption{We continue to pretrain a foundational model using records and medical knowledge sources. This process adapts the model to the data distributions of medical records, freeing us to more flexibly and directly focus downstream training on specific tasks.}
    \label{tab:pretraining}
\end{table}

\subsection{Instruction Fine-Tuning}
Any task that we use to process records can be performed by either a human abstractor using software developed internally or by our LLM. Data generated by abstractors is used for LLM training, as well as model auditing, label correction, and inter-rater reliability studies against clinicians. 

We categorize tasks into two types: \emph{structuring} and \emph{abstraction}:

\begin{figure*}[t]
\centering
\includegraphics[width=0.85\textwidth]{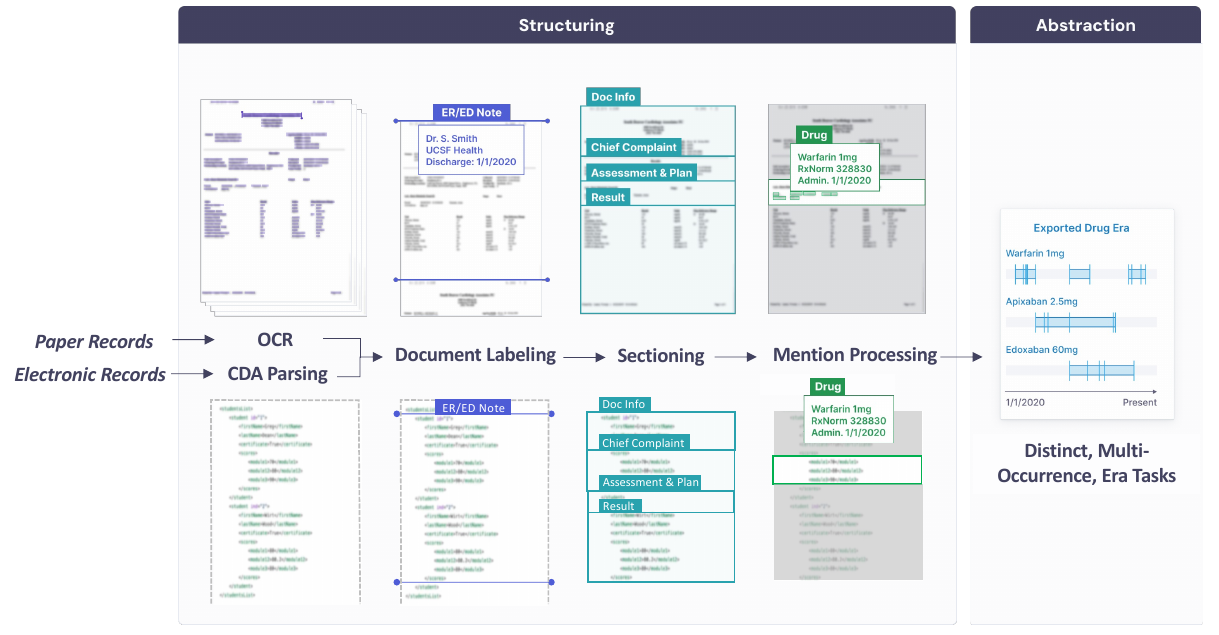}
\caption{The workflow we implement to structure and abstract records sourced electronically and on paper.}
\label{fig:hipe}
\end{figure*}

\subsubsection{Structuring Tasks}
\label{sec:structuring}
Once retrieved, the challenge of turning medical records into usable, reliable data is daunting~\cite{Tayefi2021Challenges, Rajkomar2018Scalable}. Issues found in their pages include contradictions, errors, omissions, and even notoriously difficult-to-read handwriting for paper records. Pervasive problems like the misdiagnoses in Section~\ref{sec:intro} can happen for reasons as mundane as a provider choosing the wrong option from a drop down list in EHR software. Or, a Medication List section that is intended to be a universal source of truth may combine patient recollections with provider sourced data. Even aligning data representing the same real-world concept is a challenge, and though ontologies such as SNOMED, RxNorm, and the ICD standard are intended as solutions, coding standards change over time, and differ across facilities and providers~\cite{Lee2014Literature, Nelson2011Normalized, Quan2005Coding}. 

For all of these reasons, getting data out of records and into a structured form suitable for analysis -- in our case, modeled after the OMOP Common Data Model -- requires more than just digitization for paper records or parsing for electronic records. Our \emph{structuring} tasks implement the following steps (Figure~\ref{fig:hipe}):

\begin{itemize}
\setlength{\itemsep}{3pt}
\item \textbf{OCR/CDA Parsing -} Accurate text is critical to all downstream processing. For paper records, we apply an OCR model trained on 6.2B words and bounding boxes, capturing the layouts, styles, artifacts, and language common in our records. For electronic records, we parse data in the Clinical Document Architecture (CDA) format to access text directly~\cite{dolin2001hl7}. 
\item \textbf{Document Labeling -} We tag all records with document-level metadata. This includes document boundaries, e.g. since facilities often consolidate many visits into a single paper record, as well as attributes like visit length and type, provider identity and specialty, and facility names.
\item \textbf{Sectioning -} We subdivide documents into sections, which capture data provenance. This step includes finding section boundaries based on arbitrary contents and, for paper records, physical layouts. We code types such as \emph{History of Present Illness}, \emph{Chief Complaint}, \emph{Medication List}, etc.
\item \textbf{Mentions-Processing -} Similar to named entity recognition (NER)~\cite{Chen2015A}, we identify clinical concepts, their attributes, and the relationships that link them. We do this for medications, lab tests, vital signs, procedures, and conditions. Example attributes include reference ranges for labs, doses for medications, etc. We then align the resulting \emph{entity mentions} to ontologies appropriate for their domain.  
\end{itemize}

\subsubsection{Abstraction Tasks}
\label{sec:abstraction}
While structured data reflects what is \emph{written}, our abstracted data represents the clinical view of a patient's medical history. For example, a structuring task will recognize and normalize every mention of a drug, but those named-entities alone are often not enough to confidently say when a patient started the drug, when they stopped, and why~\cite{Uzuner2010Extracting}. This is apparent in Figure~\ref{fig:context_example}, which shows snippets used to abstract the treatment course for a medication. We see that the most recent note captures a discussion of bladder control difficulties alongside two medications. It definitively notes the date that the drug was stopped, which is confirmed by the prior note. However, only the oldest note calls out worsening side effects to give us the stop reason, while leaving unclear the date of final administration. In this example, all three notes must be examined together to model the treatment course -- no \emph{one} note nor its structured data tells the full story -- and this need motivates abstraction.

New abstraction tasks are defined by configuring the desired \emph{output}, the \emph{input} source material, and a \emph{protocol} to follow. The output is a target concept, e.g. a drug code, and one of three data types:

\begin{itemize}
\setlength{\itemsep}{3pt}
\item \textbf{Distinct} variables, such as a primary diagnosis.
\item \textbf{Multi-occurrence} variables, such as episodic pain crises or clinical relapses.
\item \textbf{Era} variables that capture spans of time associated with a clinical status, such as when a patient was on a medication.
\end{itemize} 
The inputs define what context from the patient's full set of records to consider when completing the task. Inputs are configured based on document and visit metadata, such as document type, date, and provider specialty, as well as search hits for concepts related to the output. The protocol consists of definitions, guidelines, and examples to mold clinical expertise into a rigorous, repeatable process. They are designed collaboratively by clinicians and researchers, and can include multiple rounds of training with abstractors, assessment, feedback, and revision. 

In our prior example, drug era abstraction would be instantiated with the drug's associated RxNorm code and an \texttt{era} datatype; source material would be configured to look to Progress Note sections from neurology documents, as well as search hits for any drug associated with Multiple Sclerosis (MS); and the protocol would provide guidelines for navigating ambiguities, such as contradictions between primary care and specialist providers.

Our approach to designing abstraction tasks started from the observation that clinicians intuitively abstract medical information as they read records. Through user study, we discovered that the key to abstracting a nuanced treatment course lay in supplementing structured data with a provider's clinical knowledge and provenance information to contextualize and filter what was written. \emph{In this way, abstraction tasks enable us to train an LLM to mimic clinicians}: when a new task is launched to abstractors at scale their outputs become the labels for training, the configuration of input source material is the basis for LLM context generation, and the abstraction protocols provided to abstractors become the starting point for task prompts.

\begin{figure*}[t]
\begin{lstlisting}[breaklines, basicstyle=\tiny,frame=single]
### Snippets
7293 - Progress Note on 2019-12-02 Neurology, found concepts [drug_1]:
   HISTORY OF PRESENT ILLNESS:
   first_name was last seen on 5/31/2019 first_name has decided that since starting the DMT, she has been on a "downhill slide". She has received 5 doses first_name reports feeling worse after each dose and is overwhelmed by the side effects. She presents today using her walker to assist with

7296 - Progress Note on 2020-12-09 Neurology, found concepts [drug_1]:
   HISTORY OF PRESENT ILLNESS:
   first_name was last seen on 06/02/2020 first_name_1 her most recent dose of drug_1 was on 05/31/2019, she has been off drug_1 since that time. first_name feels that with oxybutynin her bladder control is better at night, but she still experiences frequency (probably

7305 - Progress Note on 2021-06-30 Neurology, found concepts [drug_1]:
   first_name was last seen 6/02/2020. She has been off drug_1 since 5/31/19, when she received her most recent dose of drug_1. first_name has some bladder urgency first_name wears incontinence pads daily. She takes oxybutynin ER 15mg before bed but still wakes 1-2 times per night to void. She continues to use stool softeners to help her move her bowels daily.
\end{lstlisting}

\caption{Neurology notes provided as context to a drug era task that includes \emph{stop reason}. We first turn to notes from specialist physicians for the most definitive account of the disease. Here, the most recent two notes co-mingle a discussion of worsening bladder control with a definitive date the patient stopped the drug. The oldest note makes a link between consistent worsening feelings and the drug, which provides the stop reason. Only with these three notes together can we piece together the era end date and stop reason.}
\label{fig:context_example}
\end{figure*}

\subsubsection{Fine-Tuning Dataset}
Table~\ref{tab:training_mix} further categorizes the 86 task types used to fine tune \ourmodel today. Figure~\ref{fig:simple_task} provides an example prompt for drug structuring, while Figure~\ref{fig:complex_task} provides an example for drug era abstraction. Each task is paired with labels collected from our corpus.

Many tasks look longitudinally across several records. On average, patient data spans 10 years of care, and 20 years at the 90th percentile. We use disease specific \emph{completeness} definitions to determine when our data fully documents a patient's primary condition over time, e.g. at-least one neurology office visit per 18 months for MS. Today, our data contains 5 years of complete documentation per patient on average. 

Our labels capture a large degree of stylistic and content variation. Records have been sourced from 100k+ care sites in aggregate, capturing an array of different documentation tools, individual writing styles, specialties, diseases, etc. They have been both structured and abstracted for research questions associated with 60+ study datasets. On average, patients have records from 10 different facilities, though some have data from as many as 140 covering their course of treatment. 

Data volume is substantial. Paper records average 30 pages, though we have retrieved records as long as 24,000 pages for heavy users of the healthcare system with complex diseases. We electronically retrieve $290$ files per patient on average, and for image data, 5,000 image slices per patient. Our full dataset contains labels collected over approximately 10 years by a workforce of several thousand abstractors in sum, and totals more than 350M labels.

\begin{table}
    \centering
    \begin{tabular}{lcc}
         \textbf{Capability}
         &  \textbf{\begin{tabular}[c]{@{}c@{}}\# Task \\ Types\end{tabular}}& \textbf{\begin{tabular}[c]{@{}c@{}}\% Training \\ Samples\end{tabular}}\\
    \cline{1-3}
    \cline{1-3}
         \multicolumn{1}{l|}{General Medical Knowledge}&  17& 29\%\\
         \multicolumn{1}{l|}{Structuring - Metadata \& Provenance}&  14& 3\%\\
         \multicolumn{1}{l|}{Structuring - Conditions}&  18& 31\%\\
         \multicolumn{1}{l|}{Structuring - Medications}&  12& 9\%\\
         \multicolumn{1}{l|}{Structuring - Labs}&  7& 4\%\\
         \multicolumn{1}{l|}{Structuring - Vital Signs}&  6& 7\%\\
         \multicolumn{1}{l|}{Abstraction - Conditions}&  4& 4\%\\
         \multicolumn{1}{l|}{Abstraction - Medications}&  8& 13\%\\
    \end{tabular}
    \caption{For each stage of our structuring and abstraction processes, we perform multiple tasks in sequence, e.g. finding vital sign names, and then attributes for each vital sign in a second pass.}
    \label{tab:training_mix}
\end{table}

\subsubsection{Task Decomposition and Context Generation}

The categories listed in Table~\ref{tab:training_mix} each contain multiple task templates. These arise in part when we decompose complex outputs into easier tasks that form dependency chains. For example, with vital signs we use one task to first identify all vital sign names within a section, another to find attributes given a name, and finally a last task to normalize units and align names to an ontology. This decomposition injects a helpful inductive bias between the measurement name and its attributes -- given ``Body Weight", the model is more likely to latch onto numbers in typical ranges for pounds and kilograms than those for degrees Farenheit. Though simple, such induction greatly improves tolerance to the variations and artifacts we encounter across facilities, e.g. when tables are intermingled with narrative text in unexpected ways.

Beyond exploiting instruction fine-tuning to shape inductive biases, we engineer the context of each task carefully. Similar to REALM~\cite{Guu2020REALM} and RAG~\cite{lewis2020retrieval} we build context by finding relevant snippets of records and concatenating them into our prompt. As discussed in Section~\ref{sec:abstraction}, context retrieval is configurable based on metadata and search terms, and we implement it today using ElasticSearch. We found that this system design struck the best initial tradeoff between performance and flexibility, allowing us to quickly iterate task design. In the future, we expect to explore end-to-end training~\cite{Li2022Decoupled}, though we note that it comes with more complex system interdependencies.

\begin{figure}
\begin{lstlisting}[breaklines, basicstyle=\tiny,frame=single,breakautoindent=false,breakindent=0ex]
### Section:
 IMPRESSION:
 She continues symptomatically and I'm going to reissue a prescription for prednisone 1000 mg a day for 5 days. I am hoping this will quiet down her symptomatology. If she continues with c she will contact me. We had a long discussion about what to do for her. She has decided she would like to switch from Copaxone to drug_1 and I think this is a good choice. She will have blood today for JC virus and vitamin D level. She will help the paperwork for drug_1 and I will see her in followup in January. I have asked her to call as needed.

### Instructions:
In the section above, find all unique mentions of these concepts:
 - Avonex (RxNorm 153326)
 - Copaxone (RxNorm 135779)
 - Tecfidera (RxNorm 1373484)
 - drug_1 (RxNorm CODE)
 - Zeposia (RxNorm 2288407)
 - dimethyl fumarate (RxNorm 1373478)
 - rituximab-abbs (RxNorm 2105824)
 - teriflunomide (RxNorm 1310520)

Output a list of JSON objects like:
[
    {
        "concept": "Copaxone (RxNorm 135779)",
        "other_fields": {
            ...
        },
    },
    ...
]
With these optional other fields:
 - ANY__SUBJECT (who the mention refers to)
 - DRUG__INSTRUCTION (the frequency a patient is to take a medication; often
                      daily, every other day, etc.)
 - DRUG__STOP_REASON (why the patient stopped taking the drug (a SNOMED item
                      and code))
 - DRUG__STRENGTH (the drug dosage; often in mgs or mg/mL)
\end{lstlisting}
\caption{This task implements clinical named entity recognition, one of our simpler structuring tasks.}
\label{fig:simple_task}
\end{figure}

\begin{figure}
\begin{lstlisting}[breaklines, basicstyle=\tiny,frame=single,breakautoindent=false,breakindent=0ex]
### Snippets
7493 - Progress Note on 2011-06-16 Psychiatry, found concepts [drug_1]:
   use street drugs, doesn't smoke; has caffeine in a.m.
   MEDICAL HISTORY Is on an ex perimental medication (infusions of drug_1) 
   for MS under the care of Dr Fox and is doing fairly well with his MS. 
   Has had partial

7485 - Progress Note on 2014-02-17 Neurology, found concepts [drug_1]:
   HPI:
   Mr. last_name is here for Camms EXT Visit Month 36. His last 
   drug_1 was Month 12 in 1/25/2010. Reports c/o pneumonia 
   (start 2/3/2014). Seen by PCP. Chest X-ray abnormal. Started Prednisone 50 
   mg po qd, Zithromax Z pak,
   
7498 - Radiology Report on 2014-02-20, found concepts [drug_2]:
   MRI BRAIN WITH AND WITHOUT CONTRAST: 2/20/2014
   HISTORY: Multiple sclerosis. There is no indication the patient is on drug_2             
   COMPARISON: Head CT 03/18/2011 (Seton Northwest Hospital), report, a 
      previous outside MRI has been
   
7483 - Progress Note on 2014-08-19 Neurology, found concepts [drug_1]:
   HPI:
   Mr. last_name is here for CAMMS Extension Visit Month 42. Last 
   drug_1 January 2010. No PE to be done today. No new symptoms, 
   no relapse. Informed consent given, all questions answered. Signed. Copy 
   to patient. No new neurologic



### Instructions:
Predict the expected drug eras after observing these snippets.

Use this format for "start_date" and "end_date": 
  {"date": "YYYY-MM-DD", 
   "precision": "DAY|MONTH|YEAR", 
   "specificity": "KNOWN_DATE|TRUE_DATE"}. 

Choose "TRUE_DATE" if the patient started or ended the  drug on that date. 
  Otherwise, choose "KNOWN_DATE" for the first or last recorded usage of 
  the drug.

Return a JSON list of drug eras. Each drug era should use 
the format
{
    "concept_name": "...",
    "start_date": {...},
    "end_date": {...},
    "stop_reason": "...",
}

For this task you should consider the following concepts:
- drug_1 (synonyms BRAND_NAME1)
- drug_2 (synonyms BRAND_NAME2)
\end{lstlisting}
\caption{This is an example of a drug era task for medications associated with MS.}
\label{fig:complex_task}
\end{figure}  
\section{Safety and Quality Checking}
\label{sec:qc}
Both \ourmodel's outputs and abstractor labels pass through several layers of validation before they are shown to users or included in a research dataset. This section summarizes them. 

\subsection{Uncertainty-Driven Manual Review}
A primary use of \ourmodel is to automate record processing, while maintaining the same accuracy as clinicians performing the same task. Even for clinicians, we observe that some records are far more difficult to understand than others, often for complex reasons. For example, a poor quality scan of a decades-old handwritten note likely induces more mistakes due to confusion about the text than the output of a modern, widely used EHR system. At the same time, we see that modern EHRs produce large amounts of redundant information, spreading the most important data sparsely over many pages.

For these reasons, we train secondary \emph{uncertainty models} to classify when the outputs of \ourmodel should be routed for additional manual review. Today, these models are analytic classifiers, not LLMs, and take into account features such as \ourmodel's logits, its outputs, information about input text such as OCR confidence, and document metadata. They are trained to detect when outputs fall short of gold standard labels. We audit the decisions of these uncertainty models  continually in production by randomly selecting tasks for review by abstractors -- the resulting dataset can then compared to the uncertainty-model's prediction. Should the quality of routing decisions slip, we are able to reprocess affected data, while retraining or recalibrating the uncertainty model. We remark that these models are highly accurate, though their implementation is not discussed in detail in this paper. 

Abstraction tasks for research study datasets are \emph{not} processed in a fully automated fashion today and are always routed for manual review. This is because they are often intended for use cases that require human verification to meet regulatory standards. For this subset of tasks, \ourmodel's abstraction outputs are instead treated as hypotheses that can speed  abstractors' work. For these tasks, we track \ourmodel's impact on abstractor task-time for a fixed accuracy bar, though we do not report on abstractor efficiency in this paper. 

\subsection{Rules-Based QC}
All \ourmodel and abstractor outputs are subjected to rules-based quality control (QC) for data conformance and plausibility. Conformance checks look to ensure basic correctness, e.g. that dates are valid, codes are present in an ontology, and attributes that cannot be null are indeed populated. Plausibility rules incorporate more clinical and disease-specific knowledge, for example that a drug does not start before a confirmed diagnosis date when appropriate, or that conditions only possible in females are not associated with male patients. When a rule violation is detected, it is logged with a `warn' or `error' priority level. Errors are prevented at point of entry, while warnings are routed to an escalation workflow for manual correction or suppression. Both general and disease-specific QC rules are created and continually expanded by a team of epidimiologists, clinicians, and biostatisticians. An example set of plausibility rules in production today is shown in Table~\ref{tab:qc_rules_pnh}.

\subsection{Agreement and Accuracy}
Labels assigned or verified by human abstractors are subject to additional quality checking (QC) tasks that ensure consistent performance over time and among abstractors~\cite{Pan2005Ensuring}. For abstraction task types, a second blinded task is performed based on a configurable sampling rate. In cases of disagreement, the result is adjudicated by a third abstractor. We also perform a smaller number of random audits by clinicians with a higher level of expertise than abstractors to ensure \emph{consistent} results are indeed \emph{correct}. For structuring tasks, which involve smaller units of work and less clinical judgment, QC tasks are not blinded and are performed by team members identified to be high performers. All QC sampling rates are configurable by percent of data volume, by concept, by abstractor performance level, and by study in the case of research datasets.

\begin{table}
    \centering
    \begin{tabular}{p{0.12\linewidth}| p{0.1\linewidth} | p{0.55\linewidth}}
        Cond. & Type & Plausibility Rule Description \\
        \hline\hline
        PNH & Err & PNH breakthrough hemolysis occurs within a drug era \\
        PNH & Warn & LDH collection date +/- 3 days from breakthrough hemolysis start date\\
        PNH & Warn & Acute kidney injury era has a plausible duration (> 100 days)\\
        PNH & Warn & PNH drug treatment eras should not overlap \\
        PNH & Warn & Eculizumab dose should not be less than 600 mg\\
    \end{tabular}
    \caption{A subset of plausibility rules applied during abstraction tasks for patients diagnosed with Paroxysmal Nocturnal Hemoglobinuria (PNH). These rules are continually expanded by a team of epidemiologists, clinicians, and biostatisticians.}
    \label{tab:qc_rules_pnh}
\end{table}
\section{\ourmodel Training \& Evaluation}
\label{sec:eval}
This section presents results from a small version of our LLM built from Meta's Llama3.1-8B foundational model. We continue pre-training it with one pass over the 28B token dataset in Table~\ref{tab:pretraining}, and then perform instruction fine-tuning using a single pass over approximately 8B tokens representing the task mix in Table~\ref{tab:training_mix}. During fine-tuning, we regularize using loss smoothing~\cite{Zhang2020Delving}, and linearly ramp loss from $0$ to $2.0e^{-5}$ over 500 steps before linearly decaying back to $0$. We evaluate the performance of \ourmodel-8B on medical benchmarks and production tasks, comparing it to the best general and domain-adapted models available.

\subsection{Common Benchmarks}
\label{sec:benchmarks}

\begin{figure}
\centering
\includegraphics[width=0.42\textwidth]{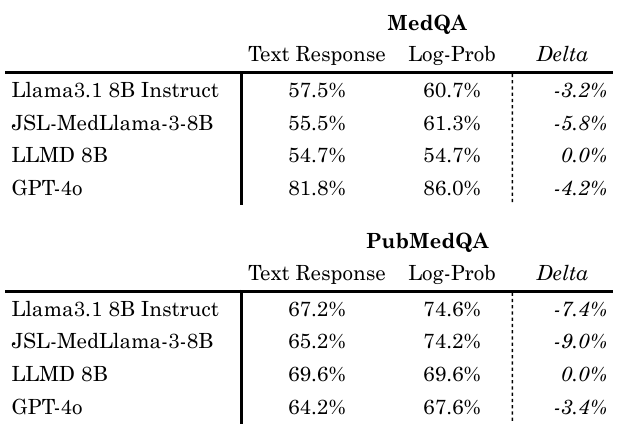}
\caption{We observe significant differences in the accuracy of 
most models when scored using ranked log-probabilities compared
to their direct text responses. LLMD is trained specifically to output correct text responses.}
\label{fig:benchmark_logprob}
\end{figure}

\begin{figure*}[htbp]
\centering
\includegraphics[width=0.95\textwidth]{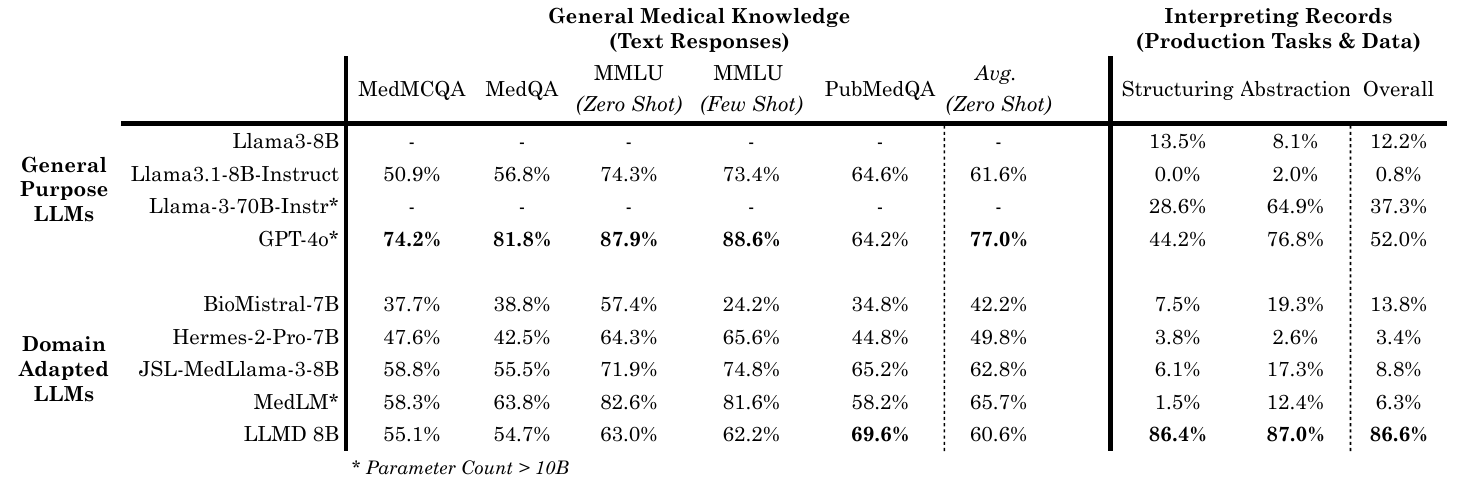}
\caption{\ourmodel achieves state of the art performance on PubMedQA text responses, and performs significantly better on the types of tasks needed to structure and abstract medical records in production.}
\label{fig:club_overall}
\end{figure*}

To analyze accuracy on medical benchmarks, we consider both log-probability scoring and text-response scoring. Given a prompt question as input, log-probability scoring ranks several possible answers in terms of the probability of emitting them as the next set of tokens; if the correct answer has highest probability relative to other options, the model is deemed correct.  This approach is most commonly reported due to the ease of comparing LLMs with different, ostensibly superficial, response styles, as well as its compatibility with legacy classification models~\cite{wang2024looktextinstructiontunedlanguage}. In contrast, text-response scoring simply feeds a question into an LLM and scores its output for correctness. This method is more difficult to report because it requires bespoke system-prompt tuning and output parsing when evaluating multiple models. 

These two methods reveal different things about LLMs. Log-probability scoring probes how well a model learns the relationships between domain concepts, while factoring out sensitivity to input and output perturbations. Text response scoring provides a more direct assessment of how an LLM might perform in production when inputs and outputs are unconstrained. A consistent relationship between the two is not guaranteed and Figure~\ref{fig:benchmark_logprob} confirms this: similar to recent work~\cite{wang2024looktextinstructiontunedlanguage}, we find that LLMs with the highest benchmark scores are far less accurate when giving text responses than their log-probabilities would suggest. In fact, no 8B parameter model met the 60\% bar in its text responses to MedQA colloquially associated with passing the US medical licensing exams, despite log-probability scoring showing that several encode the knowledge to do so. We note that \ourmodel is trained based on the quality of its text responses, minimizing the gap between scoring methods.

Focusing in on the quality of text responses, Figure~\ref{fig:club_overall} shows that \ourmodel-8B achieves state of the art responses on PubMedQA over all models, regardless of domain specialization or parameter count. This result confirms the power of continued pretraining and suggests that records themselves have content useful for improving benchmark performance. These may include examples of medical facts made manifest in patient assessments and test results, or practical explanations of knowledge in the notes of providers. 

We also notice two important behaviors in the text response scores across models. First, we find that good performance transfers less-effectively among benchmarks when scoring text responses than probability scoring has previously suggested~\cite{Leaderboard}. In fact, several models slide below the accuracy of the Llama3.1-8B-Instruct base model  despite strong performance on one or two benchmarks. Second, we see that general models with large parameter counts routinely outperform domain models: on the MedQA benchmark, llama3.1 Instruct has the best performance among 8B parameter models, while GPT-4o bests MedLM, the most advanced extension of the med-palm2 family.

These results support our experience that performance on medical knowledge benchmarks rarely determines the effectiveness of an LLM when working with records in production. Even on questions probing the same domain that have been curated of phrased differently, the tolerance to variations inherent in large production-grade general models is more important than medical knowledge. In the next section, when we evaluate performance on medical records with a much higher degree of noise and variation, this effect is even more pronounced.

\subsection{Production Workload Accuracy}
\label{sec:club_eval}

\begin{figure*}[t]
\centering
\includegraphics[width=0.95\textwidth]{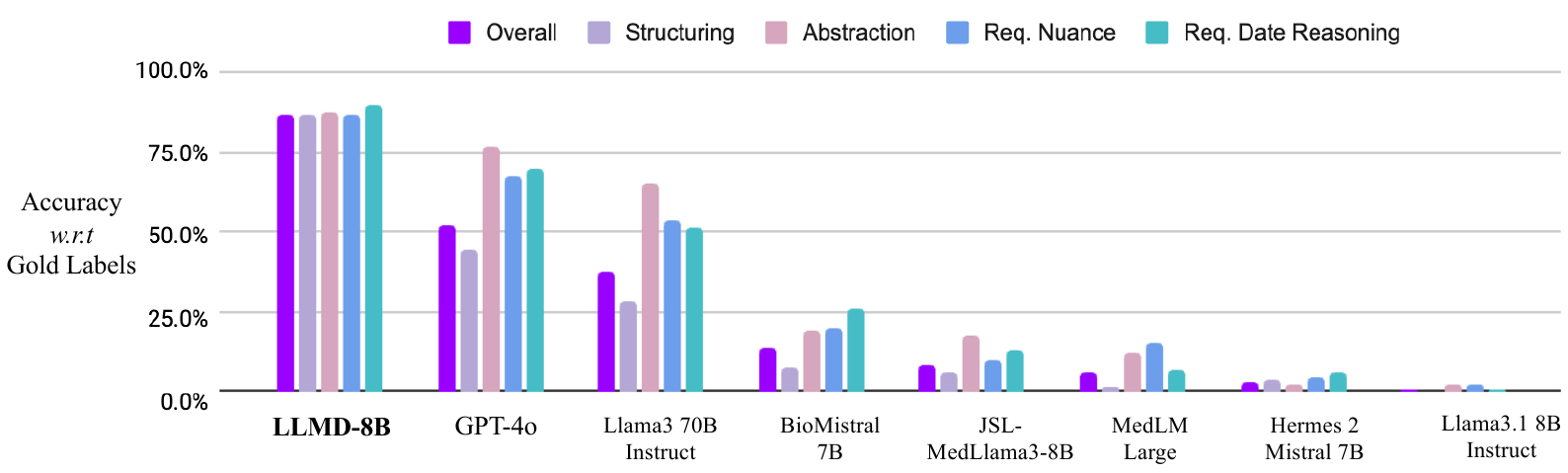}
\caption{\ourmodel learns intricate patterns in medical records, including the nuances to resolve conflicting information or to navigate workflow information embedded in records.}
\label{fig:club_nuance}
\end{figure*}

Our production tasks allow us to analyze the strengths and weaknesses of LLMs on real-world records drawn from a broad, representative patient population. Results in this section report accuracy against gold labels assigned and checked by abstractors. 

The comparison in Figures~\ref{fig:club_overall} and~\ref{fig:club_nuance} breaks model performance out by structuring and abstraction, and provides an overall score reflecting the task mix in Table~\ref{tab:training_mix}.
It shows that \ourmodel-8B handily beats comparison models, reflecting the importance of fine-tuning on tasks with labels when analyzing records. Consistent with the results in Section~\ref{sec:benchmarks}, the next best performers are large general-purpose LLMs. Our data further shows that the gap between these and models focused on domain knowledge is substantial on real-world patient data.

Examining individual responses, we observe that \ourmodel-8B leverages both the pretraining and fine-tuning datasets to improve accuracy. For example, we see structuring tasks appropriately biased towards more plausible answers. In one representative structuring task, the domain-specific JSL-MedLlama-3 model identifies a patient's height value correctly as ``6" but improperly assigns the ``inches" unit, a choice in the LLM's context window that is \emph{close} -- inches are a valid unit for height -- but implausible given the patient age, also included as input. \ourmodel-8B does not make this mistake and correctly outputs ``feet" as the unit. Overall we observe a substantially lower incidence of implausible results in the outputs of \ourmodel-8B, which we attribute to training datasets that capture many examples of measurements for people in various states of health, at various ages, etc. 

Another class of data issue that we find \ourmodel-8B handles better than alternative LLMs involves the manipulation of lab test codes and medication identifiers. We observe a significantly higher rate of incorrect codes with other LLMs, both when they are transcribed from inputs or recalled from LLM knowledge. Moreover, we observe that the most powerful models like GPT-4o often produce hallucinated codes in plausible formats, whereas much smaller models produce non-conforming outputs that are easier to detect. This complicates quality checking, and we earmark this effect -- \emph{that a little knowledge can be a dangerous thing for LLMs} -- for further study, while noting that multi-layered validation and consistency checking is necessary for safe deployment of today's LLMs. 

Figure~\ref{fig:club_nuance} also annotates some tasks based on whether they require nuanced reasoning germane to medical records. We include tasks in this category that are interpretive in nature, such as those requiring disease-specific adjudication of conflicting information. In spot checks, we see \ourmodel-8B shine on these tasks, for example properly resolving the status of a medication found in a Medication List that was also listed as stopped in a Progress Note from the same day. Large LLMs like GPT-4o and Llama-3-70B also perform well given their ability to consistently latch onto \emph{plausible} answers, though \ourmodel wins by more often finding \emph{correct} answers. 

The last category we analyze -- tasks that require date reasoning -- demonstrates how mistakes on mundane-seeming metadata can lead to poor application-level behaviors. In many failure cases, we saw comparison models confused by the meaning of dates in medical records.  When looking into the records themselves, we found dates and times documenting facility workflows, such as when notes were written, amended, signed, or when test samples were sent off to a lab, returned, etc. Answering straightforward questions about medical histories at the application level requires disentangling this timing information. Again we see in Figure~\ref{fig:club_nuance} that direct training on example data produces the best model, and also call out this as \emph{a case where the type of medical knowledge reflected in common benchmarks is little help getting basic, fundamental questions about a patient right}.

\subsection{Long Tail Performance}
Finally, we report \ourmodel's performance on structuring two specific sets of labs: the top-100 most-common and 100 tests deemed by our clinical team to be both rare and clinically important, which we refer to as long-tail labs. An example of this latter set are measurements associated with the marker panel administered to patients diagnosed with PNH (Table~\ref{tab:pnh_labs}). Given the disease's incidence of less than ten per million people, the frequency of these tests is very low in most data samples, but their importance high. 

\begin{table}
\begin{tabular}{ll}
\multicolumn{1}{c}{Measurement Name} & \multicolumn{1}{c}{\begin{tabular}[c]{@{}c@{}}Abstracted \\ Variable Type\end{tabular}} \\ \hline
\multicolumn{1}{l|}{Type 2 RBC clone size} & Occurrence \\
\multicolumn{1}{l|}{Monocyte clone size} & Occurrence \\
\multicolumn{1}{l|}{Granulocyte clone size} & Occurrence \\
\multicolumn{1}{l|}{Total RBC clone size} & Occurrence  \\
\multicolumn{1}{l|}{Type 3 RBC clone size} & Occurrence \\
\multicolumn{1}{l|}{\begin{tabular}[c]{@{}l@{}}Lactate dehydrogenase \\ {[}Enzymatic activity/volume{]} \\ in Serum or Plasma\end{tabular}} & Occurrence                         \end{tabular}
    \caption{A PNH Marker Panel provides an example of infrequent but important measurements. These tests are performed once when a patient is diagnosed; PNH occurs in fewer than 10 in 1M people.}
    \label{tab:pnh_labs}
\end{table}

In data audited over the course of April, 2024, we find precision and recall on our top 100 labs strictly above those computed from agreement studies between two abstractors performing manual abstraction. This indicates that \ourmodel's outputs after validation are as-good or better than a trained human abstractor. Among the set of long tail labs, we find that 60 of the 100 appeared more than 10 times in our audit sample -- of these, 85\% had an F1 score above 0.80, suggesting that performance in the long tail is good, but not guaranteed. In practice, when we detect this, we are able to flag sections for patients with the associated disease for manual review by abstractors, implement QC rules to ensure we find expected measurements, and ultimately retrain \ourmodel.

We have experimented both with upsampling and data augmentation to shore up long tail concepts, and for both methods find that \ourmodel responds smoothly. We find these dynamics supportive of our claim that a large labeled dataset is absolutely critical to good performance: precision and recall on these obscure concepts are \emph{not} a given, but we do see that LLMs are well-behaved enough that model blindspots are discoverable and addressable. Our results also highlight how important the input of clinicians is, and suggests that disease-by-disease rollout is likely to produce incremental generalization for medical LLMs. 
\section{Conclusion}
This paper presented \ourmodel, an LLM capable of analyzing patient health from data available today. Central to \ourmodel's success is the finding is that for medical LLMs, training on real-world data is \emph{necessary}: even the most knowledgeable models struggle when working with medical records, and dealing effectively with messy, idiosyncratic data is the limiting factor when building medical LLMs for the real-world. 

Beyond top-line accuracy, real-world medical LLMs must perform well on data that is important and potentially under represented in training datasets. We find ample evidence that guard rail and validation system design is critical for even the most powerful LLMs known. We also find that approaches for assessing and improving performance on long-tail data are a critical issue. For \ourmodel, we address this through disease specific analysis and systems that help automate feedback from clinicians. Above all, for future medical LLMs to consistently progress, there is a need for more representative training and benchmarking datasets. 

But, while these problems are difficult, they are tractable and the results compelling. We showed \ourmodel can operate at human-level accuracy and be used to improve patient care \emph{today}. User feedback has demonstrated patients discovering new things about their health history, advocating for the highest standards of care for themselves, and making better use of precious time with their doctors. Researchers are working the same underlying data, contributed by willing patients who are highly motivated to improve treatment options for themselves and others. To date, this has produced 60+ datasets covering 50+ rare diseases, and has been the basis for compelling evidence submitted to the FDA.

\bibliographystyle{ieeetr}
\bibliography{references}
\end{document}